\documentclass[conference]{IEEEtran}
\IEEEoverridecommandlockouts
\usepackage{cite}
\usepackage{amsmath,amssymb,amsfonts}
\usepackage{algorithm}
\usepackage{algorithmic}
\usepackage{graphicx}
\usepackage{textcomp}
\usepackage{xcolor}
\usepackage{algorithm}
\usepackage{array}
\usepackage{stfloats}
\usepackage{url}
\usepackage{booktabs}
\usepackage{verbatim}
\usepackage{multicol}
\usepackage{array}
\usepackage{stfloats}
\usepackage{lipsum}
\usepackage{tikz,hyperref}
\hypersetup{colorlinks,linkcolor={blue},citecolor={blue},urlcolor={blue}} 
\usepackage{makecell}
\usepackage{pdflscape}
\usepackage{multirow}
\usepackage{etoolbox}
\usepackage{enumitem}
\usepackage{balance}
\usepackage{bbding}
\usepackage{pifont}
\usepackage{gensymb}
\usepackage{mathtools}
\usepackage{svg}
\usepackage{bm}

\def\BibTeX{{\rm B\kern-.05em{\sc i\kern-.025em b}\kern-.08em
    T\kern-.1667em\lower.7ex\hbox{E}\kern-.125emX}}
\begin{document}

\title{SwasthLLM: a Unified Cross-Lingual, Multi-Task, and Meta-Learning Zero-Shot Framework for Medical Diagnosis Using Contrastive Representations
}

\author{\IEEEauthorblockN{ Ayan Sar}
\IEEEauthorblockA{\textit{School of Computer Science} \\
\textit{University of Petroleum and Energy Studies (UPES)}\\
Dehradun, 248007, Uttarakhand, India \\
ayan.sarbwn@gmail.com}
\and
\IEEEauthorblockN{ Pranav Singh Puri}
\IEEEauthorblockA{\textit{School of Computer Science} \\
\textit{University of Petroleum and Energy Studies (UPES)}\\
Dehradun, 248007, Uttarakhand, India \\
pranavpuri08@gmail.com}
\and
\IEEEauthorblockN{ Sumit Aich}
\IEEEauthorblockA{\textit{School of Computer Science} \\
\textit{University of Petroleum and Energy Studies (UPES)}\\
Dehradun, 248007, Uttarakhand, India \\
sumit9837aich@gmail.com}
\and
\IEEEauthorblockN{ Tanupriya Choudhury}
\IEEEauthorblockA{\textit{School of Computer Science} \\
\textit{University of Petroleum and Energy Studies (UPES)}\\
Dehradun, 248007, Uttarakhand, India \\
tanupriya@ddn.upes.ac.in}
\and
\IEEEauthorblockN{ Abhijit Kumar}
\IEEEauthorblockA{\textit{School of Computer Science} \\
\textit{University of Petroleum and Energy Studies (UPES)}\\
Dehradun, 248007, Uttarakhand, India \\
abhijit.kumar@ddn.upes.ac.in}
}

\maketitle

\begin{abstract}
In multilingual healthcare environments, automatic disease diagnosis from clinical text remains a challenging task due to the scarcity of annotated medical data in low-resource languages and the linguistic variability across populations. This paper proposes SwasthLLM, a unified, zero-shot, cross-lingual, and multi-task learning framework for medical diagnosis that operates effectively across English, Hindi, and Bengali without requiring language-specific fine-tuning. At its core, SwasthLLM leverages the multilingual XLM-RoBERTa encoder augmented with a language-aware attention mechanism and a disease classification head, enabling the model to extract medically relevant information regardless of the language structure. To align semantic representations across languages, a Siamese contrastive learning module is introduced, ensuring that equivalent medical texts in different languages produce similar embeddings. Further, a translation consistency module and a contrastive projection head reinforce language-invariant representation learning. SwasthLLM is trained using a multi-task learning strategy, jointly optimizing disease classification, translation alignment, and contrastive learning objectives. Additionally, we employ Model-Agnostic Meta-Learning (MAML) to equip the model with rapid adaptation capabilities for unseen languages or tasks with minimal data. Our phased training pipeline emphasizes robust representation alignment before task-specific fine-tuning. Extensive evaluation shows that SwasthLLM achieves high diagnostic performance, with a test accuracy of 97.22\% and F1-Score of 97.17\% in supervised settings. Crucially, in zero-shot scenarios, it attains 92.78\% accuracy on Hindi and 73.33\% accuracy on Bengali medical text, demonstrating strong generalization in low-resource contexts. These results highlight the potential of SwasthLLM to serve as a scalable, language-agnostic solution for AI-assisted medical diagnosis in multilingual regions.
\end{abstract}


\section{Introduction}
The introduction of artificial intelligence (AI) into healthcare systems has significantly improved medical decision-making, particularly in diagnostic tasks. However, most existing AI-based diagnostic models are designed for monolingual scenarios and require extensive labeled data, making them impractical for multilingual and low-resource settings - common in many regions around the world. In countries like India, where patients and practitioners communicate in multiple languages such as English, Hindi, and Bengali, a scalable, language-agnostic medical diagnosis system is urgently needed.

Medical diagnosis from unstructured clinical text poses several challenges: linguistic diversity, domain-specific vocabulary, variations in syntax and semantics across languages, and the scarcity of labeled datasets, especially in non-English languages. Traditional translation-based approaches introduce noise, while building separate models per language is computationally expensive and data-intensive. Thus, there is a critical need for a unified framework capable of understanding and diagnosing medical conditions across multiple languages, particularly in low-resource contexts.

\begin{figure*}[t!]
    \centering
    \includegraphics[width=\textwidth]{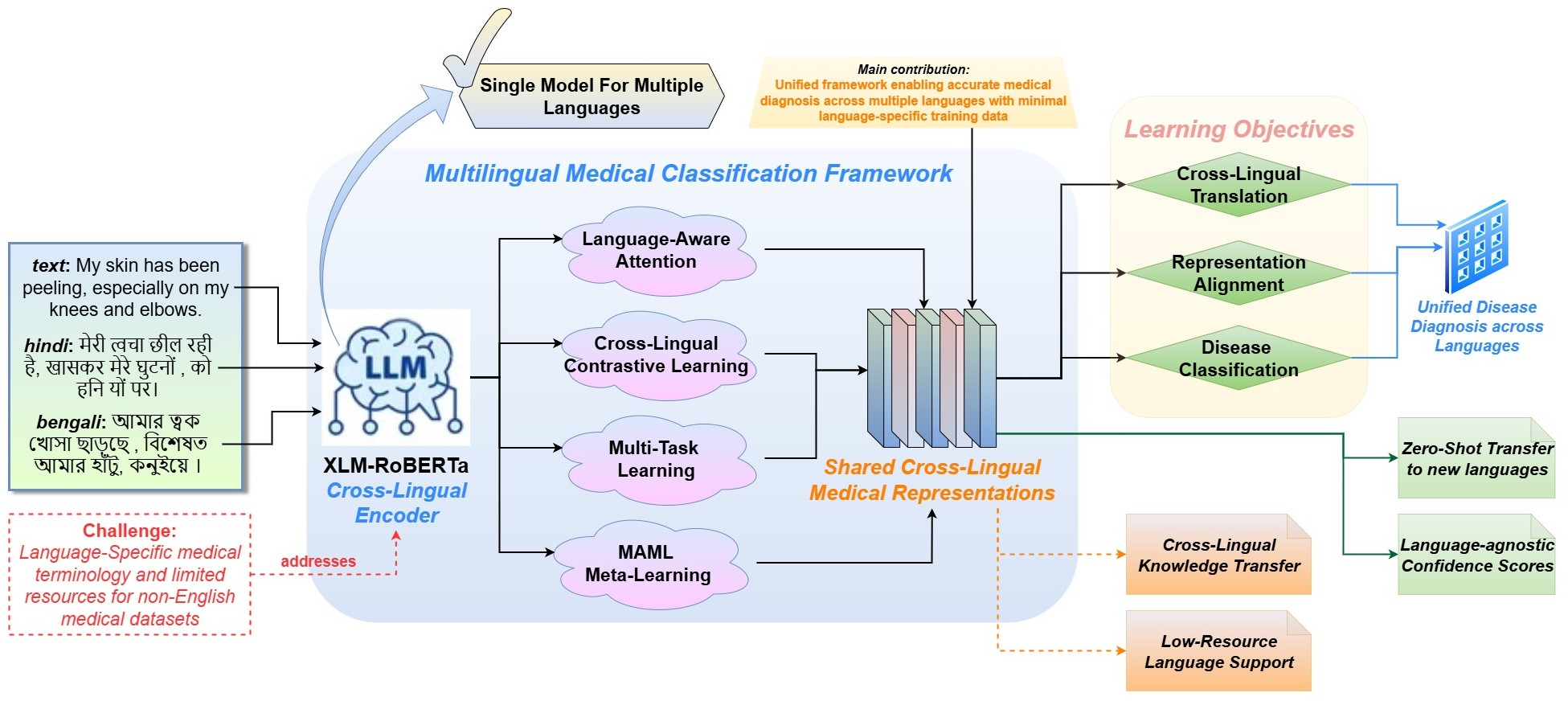}
    \caption{An overview of our proposed Multilingual Medical Classification Framework. The model utilizes a cross-lingual encoder (XLM-RoBERTa) with language-aware attention, contrastive learning, and meta-learning to create shared representations for unified disease diagnosis across multiple languages.}
    \label{fig:intro}
\end{figure*}

To address these challenges, we propose SwasthLLM, a unified cross-lingual, multi-task, and meta-learning framework for zero-shot medical diagnosis. At its core, SwasthLLM utilizes XLM-RoBERTa, a transformer-based multilingual encoder, to generate contextual embeddings from clinical text in any of the supported languages. The architecture is augmented with several key innovations, including a language-aware attention mechanism to highlight medically relevant terms, a contrastive learning module for aligning multilingual representations, a translation consistency layer, and a disease classification head for diagnostic prediction. Crucially, SwasthLLM leverages a multi-task learning paradigm, where the model jointly learns disease classification, translation alignment, and semantic embedding alignment through contrastive learning. Furthermore, it employs Model-Agnostic Meta-Learning (MAML) to enable rapid adaptation to new languages or tasks using minimal supervision, making it ideal for zero-shot and few-shot scenarios (see Fig. \ref{fig:intro}). Our key contributions are summarized as follows:

\begin{enumerate}
    \item We present a unified multilingual framework for medical diagnosis that supports cross-lingual zero-shot disease classification across English, Hindi, and Bengali.
    \item We introduced a language-aware attention mechanism that dynamically focuses on salient clinical terms relevant to diagnosis in multilingual input.
    \item We developed a Siamese contrastive learning framework to align semantic representations of medical texts across languages, enhancing zero-shot performance.
    \item We integrated meta-learning (MAML) to allow rapid adaptation to new tasks or languages, making SwasthLLM suitable for low-resource environments.
\end{enumerate}

The combination of cross-lingual transfer, contrastive representation learning, and meta-learning results in a robust and scalable diagnostic system. The rest of this manuscript is organized as follows: Section \ref{s2} reviews related work on multilingual diagnosis, cross-lingual representation learning, and meta-learning in healthcare. Section \ref{s3} details the dataset and the architecture of SwasthLLM and its components, with Section \ref{s4} detailing the experimental setup and evaluation metrics. Section \ref{s5} presents the results and analysis, with Section \ref{s6} discussing its implications and limitations, and Section \ref{s7} concludes the manuscript with directions for future work.

\section{Literature Survey} \label{s2}
Large Language Models (LLMs) have evolved rapidly from text-based systems to multimodal platforms, which enhance clinical decision support and medical imaging interpretation \cite{Yang_2024} \cite{niu2024textmultimodalityexploringevolution}. The integration of multi-task learning and meta-learning approaches has further expanded the capabilities of AI in handling complex medical data across languages and tasks \cite{Kim_2022} \cite{sharma2023multitasktrainingindomainlanguage}. With healthcare systems facing challenges such as diagnostic errors and resource constraints, the potential of LLMs to improve diagnostic accuracy and streamline workflows has significant practical and social implications \cite{mcduff2023accuratedifferentialdiagnosislarge}. For instance, recent studies report diagnostic accuracy improvements exceeding 15\% when leveraging domain-specific training and reasoning frameworks \cite{wu2023largelanguagemodelsperform} \cite{Kwon_2024}. Despite these advances, a critical problem remains in developing unified frameworks that can effectively integrate cross-lingual capabilities, multi-task learning, and meta-learning for robust medical diagnosis. Existing models often focus on single tasks or languages, which limits their generalizability and clinical utility \cite{Kim_2022} \cite{sharma2023multitasktrainingindomainlanguage} \cite{Chen_2023}. Moreover, challenges such as knowledge grounding, interpretability, and mitigating hallucinations persist, with debates on the best strategies to balance model complexity and clinical trustworthiness \cite{Savage_2024}. Some researchers emphasize the importance of the knowledge graph integration to enhance the factual accuracy \cite{yang2024kgrankenhancinglargelanguage} \cite{Gao_2023}, while others highlight the need for improved reasoning and explanation mechanisms within LLMs \cite{Kwon_2024} \cite{zhou2024interpretabledifferentialdiagnosisdualinference}. The consequences of these gaps include potential diagnostic errors and reduced adoption of AI tools in clinical practice \cite{Xie_2023} \cite{Ennab_2024}. This review involves the intersection of LLMs, multi-task learning, and meta-learning within a cross-lingual medical diagnostic context. LLMs provide advanced natural language understanding and generation capabilities \cite{Scott_2024} \cite{sharaf2023analysislargelanguagemodels}, multi-task learning enables simultaneous optimization across diverse diagnostic tasks \cite{Kim_2022} \cite{sharma2023multitasktrainingindomainlanguage}, and meta-learning facilitates rapid adaptation to new tasks and languages \cite{sharma2023multitasktrainingindomainlanguage} \cite{Chen_2023}. Together, these components form a cohesive structure aimed at enhancing diagnostic reasoning, accuracy, and applicability across heterogeneous clinical settings \cite{wu2023largelanguagemodelsperform} \cite{Kwon_2024}. The detailed review is shown in Table \ref{tab:literature} with the results and research gaps identified. By synthesizing these diverse methodologies, SwasthLLM aims to provide a comprehensive, adaptable, and equitable solution for medical diagnosis across various languages and tasks, addressing both the technical and ethical challenges of AI in healthcare.

\renewcommand{\arraystretch}{1.4}
\begin{table*}[t!]
    \centering
    \caption{Literature Survey of different methods and research gaps identified}
    \label{tab:literature}
    {\fontsize{7pt}{7pt}\selectfont \begin{tabular}{>{\centering\arraybackslash}m{2cm}|>{\centering\arraybackslash}m{3.5cm}|>{\centering\arraybackslash}m{3cm}|>{\centering\arraybackslash}m{3.5cm}|>{\centering\arraybackslash}m{3.5cm}}
        \hline
        \textbf{Paper} & \textbf{Objectives} & \textbf{Dataset} & \textbf{Results} & \textbf{Research Gaps identified} \\
        \hline \hline
         B.H.Kim et. al. 2025 \cite{kim2025largelanguagemodelsinterpretable} & Evaluate LLMs' diagnostic ability using symptom prompts, focusing on accuracy, precision, recall, and performance in general and critical tasks.& Symptom prompts from medical sources were used, but exact dataset details and prompt quantity were not disclosed.& GPT-4 had top accuracy; Gemini excelled in high-stakes tasks; others performed well in real-time or general diagnostic contexts. & Gaps include LLM bias mitigation, ethical use, and privacy compliance in healthcare diagnostic applications. \\ \hline
        G.K.Gupta et. al. 2025 \cite{gupta2024digitaldiagnosticspotentiallarge}	 & Mitigate hallucinations in LLMs by integrating curated Wikipedia-based knowledge graphs for improved factual grounding and contextual alignment.& Wikipedia-derived knowledge graphs refined to emphasize essential, context-specific data for enhancing model training and response accuracy.& Knowledge graph integration reduced hallucinations and improved factual accuracy, making LLM responses more contextually grounded and trustworthy. & Lacks analysis of KG integration challenges and long-term effects on LLM adaptability across diverse contexts. \\ \hline
        E.L. et. al. 2024 \cite{lavrinovics2024knowledgegraphslargelanguage}	& Assess LLMs’ accuracy in translating diagnostic criteria into Datalog and measure human effort needed for error correction.& Used ICD-11 CDDR diagnostic rules; method adaptable to DSM-5-TR, covering major mental health diagnostic standards.& Expert-corrected LLM outputs achieved 100\% accuracy; uncorrected LLMs misinterpreted criteria, stressing need for hybrid human-LLM systems. & LLMs need better validation for rule translation; ethical risks of patient data use demand safer, hybrid diagnostic solutions. \\ \hline
        G.R.R. et. al. 2024 \cite{rosenbaum2024medgkrpmedicalgraphknowledge} & Develop SSPEC chatbot by fine-tuning LLMs with site-specific knowledge to reduce hallucinations and enhance medical response accuracy & Used large-scale and specialized medical corpora; site-specific data collected from various hospital reception sites for fine-tuning SSPEC & SSPEC improved efficiency, communication, and empathy; achieved 99.8\% specificity and 85.0\% sensitivity in clinical safety validation & LLMs lack site-specific knowledge and clinical validation; more RCTs are needed to ensure real-world safety and effectiveness. \\ \hline
        G. Huang 2024\cite{Huang_2024} & Propose MedG-KRP to visualize medical reasoning in LLMs via knowledge graphs and assess causal understanding in medicine.& Compared 60 LLM-generated medical graphs with the BIOS biomedical KG; human medical students reviewed the outputs qualitatively.& GPT-4 excelled in human reviews but lagged in ground truth alignment; PalmyraMed showed opposite trends, revealing performance trade-offs. & Limited model diversity and subjective human reviews highlight the need for broader testing and more objective evaluation frameworks. \\ \hline
        R.K.Joshi et. al. 2024 \cite{joshi2024hallucinationsfactsenhancinglanguage}  & Improve knowledge graph learning by modeling node–relation effects and overcoming first-order subgraph limitations via RGCN and attention.& Evaluated on Cora (homogeneous) and FB15K-237 (heterogeneous) datasets, demonstrating the model's effectiveness across graph types.& Achieved improved accuracy, MRR, MR, and Hit@n; ablation studies confirmed the contribution of each model component. & Existing models inadequately capture node-relation interactions and are limited by first-order subgraph aggregation in multi-relational graphs. \\ \hline
    \end{tabular}}
\end{table*}

\section{Methodology} \label{s3}
This section outlines the key components of the proposed system, SwasthLLM, including the dataset used, the preprocessing pipeline, and the architectural design of our multilingual, multi-task learning framework.

\subsection{Dataset Overview}
\begin{figure*}[t!]
    \centering
    \includegraphics[width=\textwidth]{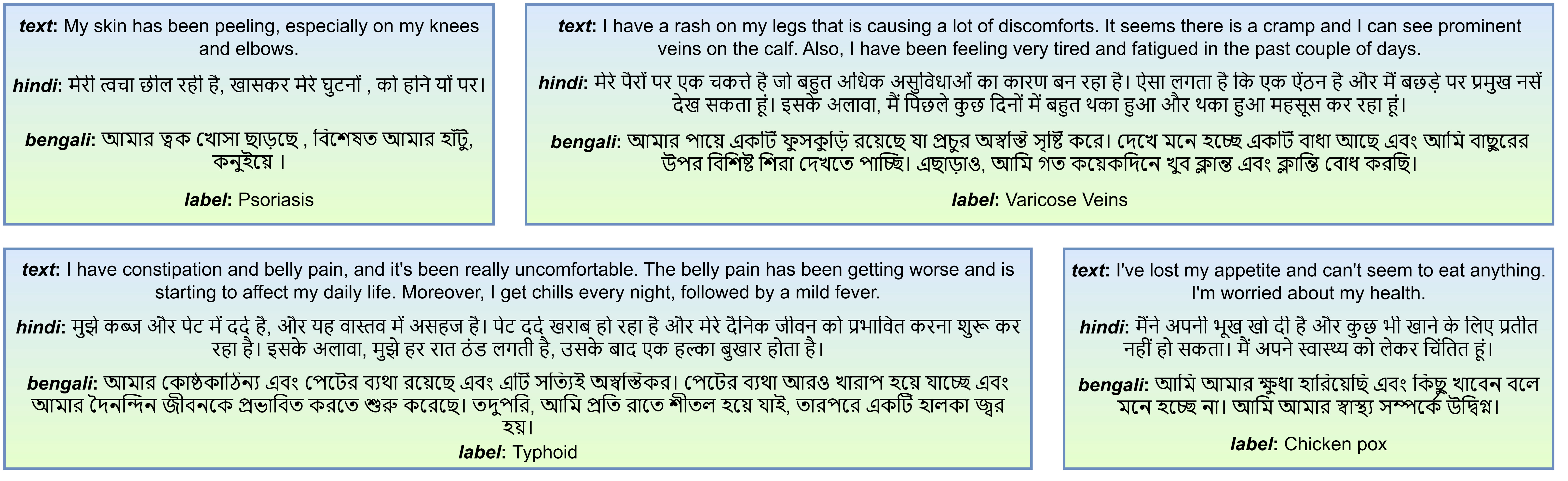}
    \caption{Structure and sample data from the Multilingual dataset.}
    \label{fig:dataset}
\end{figure*}

The dataset used in this study \cite{multilingual} is a multilingual medical text classification corpus specifically designed to support cross-lingual disease diagnosis. This consists of 1200 samples, each describing a patient's symptoms in natural language, annotated with one of the 24 distinct classes, such as Psoriasis, Malaria, Dengue, and others. What makes this dataset unique for our task is its parallel multilingual structure, where each symptom description is aligned across three languages: English, Hindi, and Bengali. For every record, the \textbf{\textit{text}} column contains the original symptom description in English, while the \textbf{\textit{hindi}} and \textbf{\textit{bengali}} columns contain direct translations of the same content in Hindi and Bengali, respectively. The \textbf{\textit{label}} column holds the target disease class associated with each symptom description (see Fig. \ref{fig:dataset}). 

The dataset is balanced, with approximately 50 samples per disease class, which helped prevent the model bias during training and enables reliable performance comparisons across all classes. This structure ensures that the dataset is well-suited not only for supervised monolingual classification but also for cross-lingual tasks, such as zero-shot diagnosis, translation consistency evaluation, and multilingual embedding alignment. Each data point thus represents a fully aligned symptom-to-diagnosis instance across three linguistic contexts, reflecting real-world clinical variability in multilingual settings and making this dataset ideal for training and evaluating models such as SwasthLLM that aim to generalize across languages with minimal supervision.


\subsection{Dataset Preprocessing}
To prepare the multilingual medical dataset for training the SwasthLLM framework, a comprehensive preprocessing pipeline was implemented to ensure consistency, language alignment, and compatibility with transformer-based architectures. First, all symptom descriptions across the three languages—English, Hindi, and Bengali—underwent basic text cleaning. This involved removing unnecessary punctuation, special characters, and excessive whitespace, followed by Unicode normalization for Hindi and Bengali to resolve encoding inconsistencies and ensure uniform script representation. Each textual input was then tokenized using the XLM-RoBERTa tokenizer, which supports multilingual subword tokenization and is compatible with the encoder used in our architecture. Special tokens such as [CLS] and [SEP] were added as per the transformer input format.

Next, the disease names in the label column were encoded into integer values ranging from 0 to 23, corresponding to the 24 unique disease classes. This mapping ensured compatibility with the softmax classification layer. Since each English sentence has an aligned translation in both Hindi and Bengali, we formed triplet-aligned entries for each sample—creating semantically equivalent parallel inputs across languages. These triplets served multiple purposes: they enabled contrastive learning by acting as positive pairs (same content in different languages), supported translation consistency learning, and provided a basis for zero-shot and few-shot cross-lingual evaluation.

The dataset was then split into training, validation, and test sets. The training set consisted primarily of English samples, ensuring that the model learns base classification performance from high-resource data. The validation set included a balanced mix of English, Hindi, and Bengali samples to tune task-specific loss weights and perform early stopping. The test set was designed to evaluate both supervised (English) and zero-shot (Hindi and Bengali) performance. Although the dataset was already balanced—with each disease class appearing approximately 50 times—class frequency distributions were monitored, and stratified sampling was used during splitting to maintain uniform label distribution across all sets. This preprocessing pipeline ensured that SwasthLLM could be trained effectively for cross-lingual medical diagnosis, translation alignment, and contrastive representation learning in a unified manner.

\begin{figure*}[t!]
    \centering
    \includegraphics[width=\textwidth]{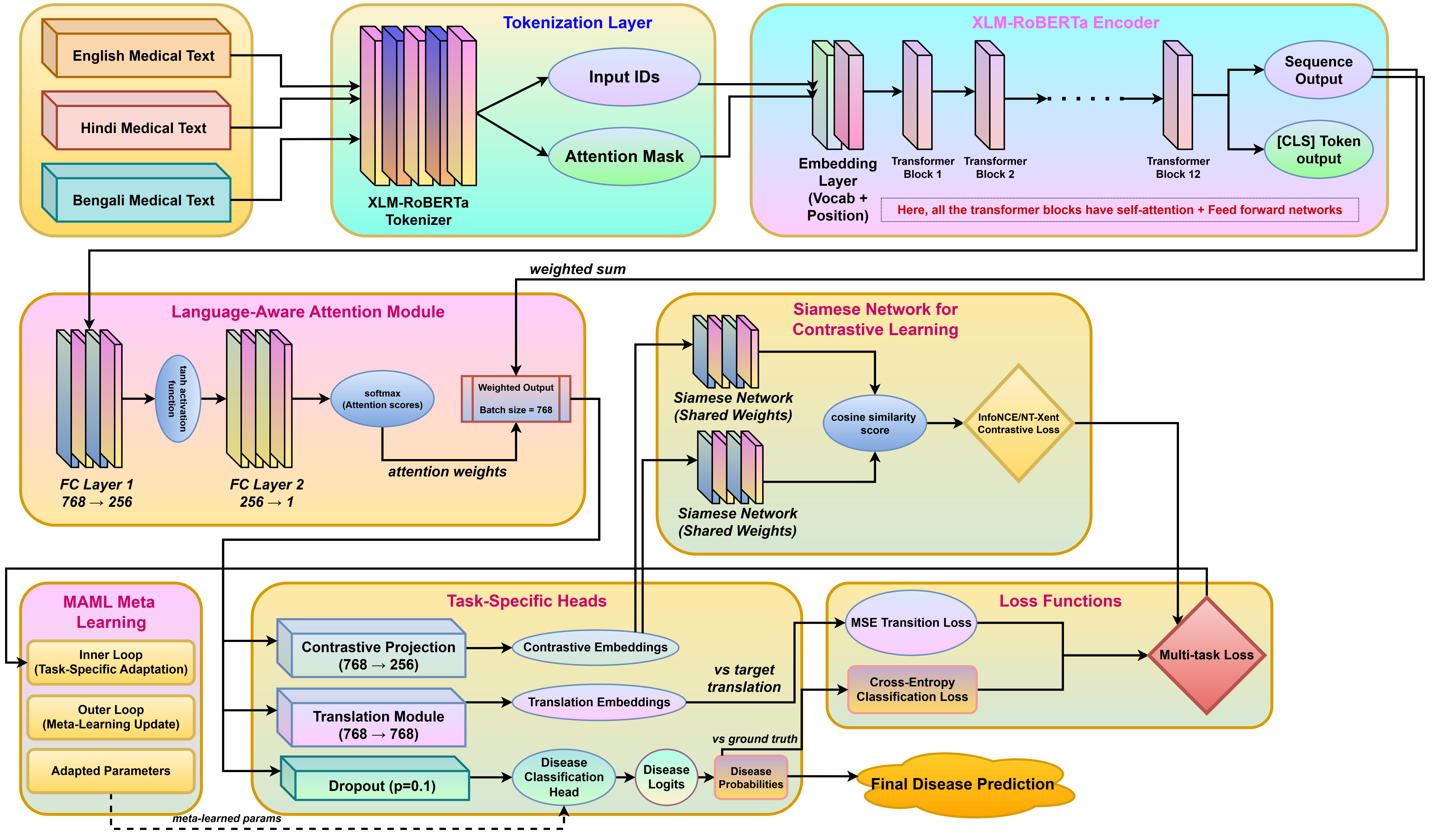}
    \caption{The architecture of the proposed framework SwasthLLM}
    \label{fig:architecture}
\end{figure*}

\begin{figure*}[t!]
    \centering
    \includegraphics[width=\textwidth]{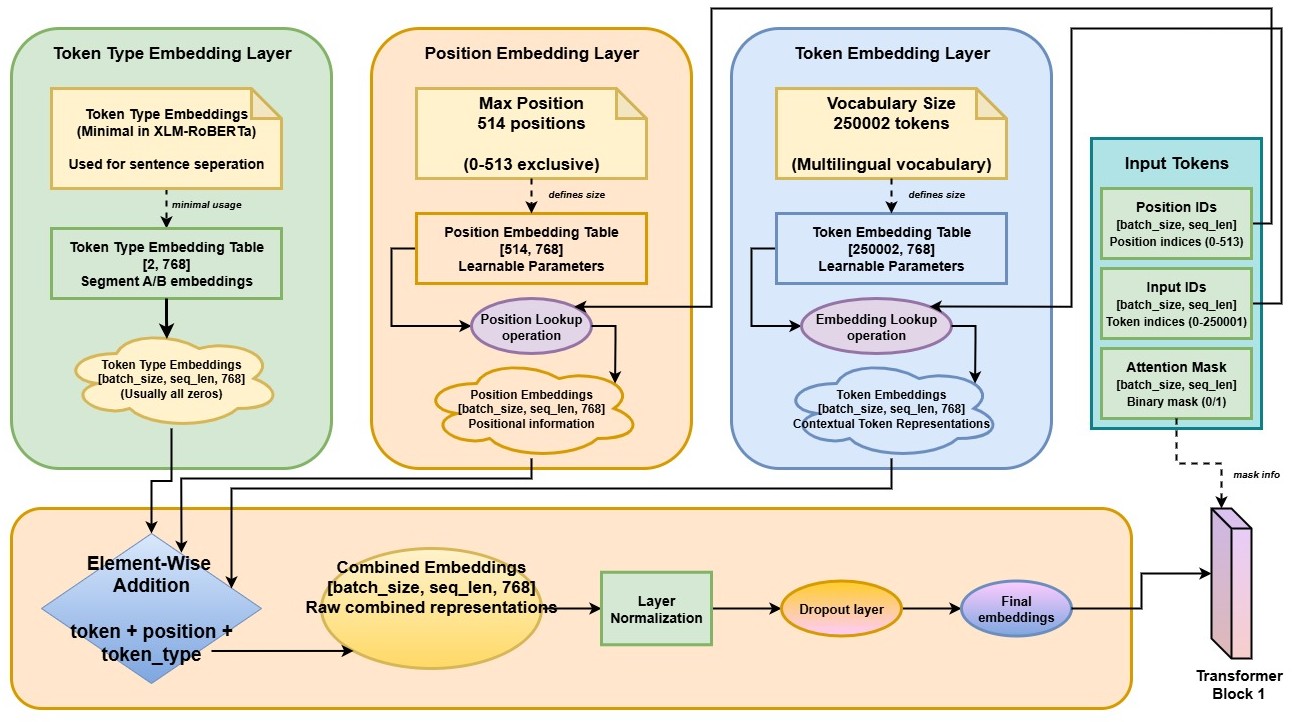}
    \caption{The embedding layer (vocab+position) in XLM-RoBERTa encoder block}
    \label{fig:embedding}
\end{figure*}

\subsection{Model Architecture}
SwasthLLM is a novel multilingual diagnostic framework designed to address the challenges of cross-lingual medical understanding through contrastive representation alignment, language-sensitive attention, and fast adaptation using meta-learning. The core of the architecture is based on the XLM-RoBERTa encoder, augmented with task-specific modules and trained using a multi-task and meta-learning pipeline. We denote a medical input sample as $x^{(l)} \in \mathcal{X}^{(l)}$, where $l \in \{en, hi, bn\}$ indicates the language (English, Hindi, Bengali). The goal is to classify the input into a disease label $y \in \mathcal{Y}$, while ensuring that embeddings $h^{(l)}$ are language-invariant and diagnostically informative.

Given an input text $x^{(l)} = \{w_1, w_2, \dots, w_n\}$, the XLM-RoBERTa encoder $\mathcal{E}_{\theta}$ produces contextual embeddings using Eq. \ref{eq1}.

\begin{equation}
    H^{(l)} = \mathcal{E}_{\theta}(x^{(l)}) = \{h_1, h_2, \dots, h_n\}, \quad h_i \in \mathbb{R}^d
    \label{eq1}
\end{equation}

The [CLS] token embedding $h^{(l)}_{[CLS]}\in \mathbb{R}^d$ is used as a compact representation of the entire input sequence. Next, we propose a language-aware attention module that dynamically assigns importance to medical terms across different linguistic structures. This mechanism enables the model to focus on critical tokens like symptoms or clinical descriptors, which may appear in varying syntactic positions in different languages. Here, the attention score for each token $h_i$ is computed as in Eq. \ref{eq2}:

\begin{equation}
    \begin{array}{c}
       e_i = v^{\top} \tan h(W_ah_i + b_a) \\
        \alpha_i = \frac{\exp (e_i)}{\sum_{j=1}^n \exp (e_j)} \\
        z = \sum_{i=1}^n \alpha_i h_i
        \end{array}
    \label{eq2}
\end{equation}

Here, $W_a \in \mathbb{R}^{d' \times d}, v \in \mathbb{R}^{d'}$, and $b_a \in \mathbb{R}^{d'}$ are learnable parameters. $z \in \mathbb{R}^{d}$ is the attended sentence-level embedding used for downstream prediction. The full breakdown of the embedding layer in the encoder is shown in Fig. \ref{fig:embedding}. This module allows the model to generalize better across languages by automatically identifying key phrases, even in morphologically rich or syntactically different languages like Hindi or Bengali. Next, a standard linear classifier takes the attended representation $z$ and predicts the disease label using Eq. \ref{eq3}.

\begin{equation}
    \hat{y} = softmax (W_cz + b_c)
    \label{eq3}
\end{equation}

Here, $W_c \in \mathbb{R}^{|\mathcal{Y}| \times d}$ and $b_c \in \mathbb{R}^{|\mathcal{Y}|}$ are trainable weights and biases. The classification loss here is defined as in Eq. \ref{eq4}.

\begin{equation}
    \mathcal{L}_{cls} = - \sum^{|\mathcal{Y}|}_{i=1} y_i \log (\hat{y}_i)
    \label{eq4}
\end{equation}

Next, to align the multilingual representations, we propose a Siamese Network structure with contrastive projection. Let $x^{(en)}$ and $x^{(hi)}$ be parallel sentences in English and Hindi describing the same medical case. The model encodes them to $h^{(en)}_{[CLS]}$ and $h^{(hi)}_{[CLS]}$, which are then projected into a shared latent space via Eq. \ref{eq5}.

\begin{equation}
    z^{(l)} = MLP_{\phi}(h^{(l)}_{[CLS]}), \quad l \in \{en, hi, bn\}
    \label{eq5}
\end{equation}

We use the NT-Xent loss (Normalized Temperature-scaled Cross Entropy Loss) for contrastive training using Eq. \ref{eq6}.

\begin{equation}
    \mathcal{L}_{contrast} = - \log \frac{exp(sim(z^{(i)}, z^{(j)})/\tau)}{\sum^{2N}_{k=1}1_{[k \neq 1]}exp(sim(z^{(i)}, z^{(k)})/\tau)}
    \label{eq6}
\end{equation}

Here, $sim(u,v) = \frac{u^{\top}v}{||u||||v||}$ is the cosine similarity, $\tau$ is the temparature parameter, and $N$ is the batch size of aligned pairs. This loss pulls together aligned sentences and pushes apart unrelated ones, yielding language-agnostic embeddings. Next, an auxiliary translation module minimizes the mean squared error (MSE) between embeddings of aligned sentences across languages using Eq. \ref{eq7}. 

\begin{equation}
    \mathcal{L}_{trans} = ||z^{(en)} - z^{(l)}||^2_2, \quad l \in \{hi, bn\}
    \label{eq7}
\end{equation}

This promotes semantic alignment, especially in cases where perfect parallelism may not exist in contrastive training alone. The total loss function combines the three components using Eq. \ref{eq8}.

\begin{equation}
    \mathcal{L}_{total} = \alpha \cdot \mathcal{L}_{cls} + \beta \cdot \mathcal{L}_{trans} + \gamma \cdot \mathcal{L}_{contrast}
    \label{eq8}
\end{equation}

Here, in the above equation, $\alpha, \beta, \gamma$ are hyperparameters that balance each task. The tuning of these will allow task prioritization based on language and data availability. Lastly, we incorporated Model-Agnostic Meta-Learning (MAML) to allow fast adaptation to new languages or diseases with minimal supervision. Let $\theta$ be the model parameters, and MAML optimizes for generalization by simulating quick adaptation over multiple tasks/languages. For the inner loop (adaptation), given a support set $\mathcal{D}_{train}^{(l)}$, we computed adapted parameters using Eq. \ref{eq9}.

\begin{equation}
    \theta' = \theta - \eta \nabla_{\theta} \mathcal{L}_{task}(\mathcal{D}_{train}^{(l)}; \theta)
    \label{eq9}
\end{equation}

For the outer loop (meta-optimization), we update $\theta$ using loss on query set $\mathcal{D}_{val}^{(l)}$ using Eq. \ref{eq10}.

\begin{equation}
    \theta \leftarrow \theta - \mu \nabla_{\theta} \mathcal{L}_{task}(\mathcal{D}_{val}^{(l)}; \theta')
    \label{eq10}
\end{equation}

\begin{algorithm}[t!]
\caption{SwasthLLM: Multilingual Medical Classification}
\label{alg:mmc}
\begin{algorithmic}[1]
\REQUIRE $\mathcal{D} = \{(x_i^{(l)}, y_i)\}$, $\mathcal{L}$, params $\alpha,\beta,\gamma,\tau,\eta$
\ENSURE Trained model $\Theta^*$
\STATE \textbf{Init:} Encoder $f_\theta$, Attn $A_\phi$, Heads $h_{cls}, h_{trans}, h_{contr}$, Meta $\Theta$
\STATE \textbf{Phase 1: Contrastive Pretraining}
\FOR{$epoch=1$ to $E_1$}
\FOR{$x$ in batch}
\STATE $H = f_\theta(x)$; $\alpha = \text{softmax}(A_\phi(H))$; $h = \sum \alpha_i H_i$; $z = h_{contr}(h)$
\STATE $\mathcal{L}_{contr} = \text{InfoNCE}(z)$; Update $\Theta$
\ENDFOR \ENDFOR
\STATE \textbf{Phase 2: Classification}
\FOR{$epoch=1$ to $E_2$}
\FOR{$x, y$ in batch}
\STATE $H \rightarrow h \rightarrow \hat{y} = \text{softmax}(h_{cls}(h))$
\STATE $\mathcal{L}_{cls} = -y \log \hat{y}$; Update $\Theta$
\ENDFOR \ENDFOR
\STATE \textbf{Phase 3: Multi-Task Training}
\FOR{$epoch=1$ to $E_3$}
\FOR{$x, y$ in batch}
\STATE $h, \hat{y}, z$ as above; $\hat{t} = h_{trans}(h)$
\STATE $\mathcal{L}_{total} = \alpha \mathcal{L}_{cls} + \beta \|h - \hat{t}\|^2 + \gamma \mathcal{L}_{contr}$
\STATE Update $\Theta$
\ENDFOR \ENDFOR
\STATE \textbf{Phase 4: MAML Meta-Learning}
\FOR{$meta\_epoch=1$ to $E_4$}
\FOR{task $T_k$ with $(D_{sup}, D_{qry})$}
\STATE $\Theta'_k = \Theta - \alpha_{in} \nabla \mathcal{L}_{sup}$; $\mathcal{L}_k^{meta} = \mathcal{L}_{qry}(\Theta'_k)$
\ENDFOR
\STATE $\Theta \leftarrow \Theta - \alpha_{out} \nabla \sum_k \mathcal{L}_k^{meta}$
\ENDFOR
\STATE \textbf{Inference:} $\hat{y} = \text{softmax}(h_{cls}(\sum \alpha_i f_\theta(x_i)))$
\STATE \textbf{Function:} \textsc{ContrastiveLoss}($z_1, z_2$) \RETURN $-\log \frac{\exp(\text{sim}(z_1,z_2)/\tau)}{\sum_k \exp(\text{sim}(z_1,z_k)/\tau)}$ 
\STATE \textbf{Function:} \textsc{LanguageAttention}($H$) \RETURN $\text{softmax}(\tanh(HW_1 + b_1)W_2 + b_2)$
\STATE \textbf{Function:} \textsc{VerifyAlignment}($\mathcal{D}$) \RETURN mean cosine sim of $z_1, z_2$ from $(x^{(l_1)}, x^{(l_2)})$ 
\end{algorithmic}
\end{algorithm}

Here, $\eta$ is the inner loop learning rate, and $\mu$ is the meta-learning rate. This training scheme allows SwasthLLM to quickly generalize to new languages and rare diseases, where only a few labeled examples are given. The algorithm for the proposed framework is shown in Algorithm \ref{alg:mmc}, along with the model architecture in Fig. \ref{fig:architecture}. This novel architecture enables SwasthLLM to achieve high classification accuracy with cross-lingual generalization, especially in zero-shot and few-shot scenarios, making it an ideal solution for multilingual medical AI applications. 

\renewcommand{\arraystretch}{1.4}
\begin{table}[t]
\centering
\caption{SwasthLLM Model Parameters}
{\fontsize{7pt}{7pt}\selectfont \begin{tabular}{>{\centering\arraybackslash}m{3cm}|>{\centering\arraybackslash}m{2.6cm}|>{\centering\arraybackslash}m{1.9cm}}
\hline
\textbf{Component} & \textbf{Parameter} & \textbf{Value / Description} \\
\hline
\multirow{5}{*}{\textbf{Encoder}} 
    & Base Model & XLM-RoBERTa-base \\ \cline{2-3}
    & Embedding Dimension ($d$) & 768 \\ \cline{2-3}
    & Max Sequence Length & 128 tokens \\ \cline{2-3}
    & Vocabulary Size & 250,002 \\ \cline{2-3}
    & Dropout (encoder) & 0.1 \\
\hline
\multirow{4}{*}{\textbf{Language-Aware Attention}} 
    & Hidden Dimension ($d'$) & 256 \\ \cline{2-3}
    & Activation Function & Tanh \\ \cline{2-3}
    & Output Units & Scalar attention scores per token \\ \cline{2-3}
    & Dropout (attention layer) & 0.1 \\
\hline
\multirow{3}{*}{\textbf{Disease Classification Head}} 
    & Number of Layers & 1 (Linear) \\ \cline{2-3}
    & Output Units & Number of disease classes ($|\mathcal{Y}|$) \\ \cline{2-3}
    & Activation & Softmax \\
\hline
\multirow{4}{*}{\textbf{Contrastive Projection Head}} 
    & Projection Layers & 2 (Linear + ReLU + Linear) \\ \cline{2-3}
    & Projection Dimension & 128 \\ \cline{2-3}
    & Loss Function & NT-Xent (InfoNCE) \\ \cline{2-3}
    & Temperature Parameter ($\tau$) & 0.07 \\
\hline
\multirow{2}{*}{\textbf{Translation Module}} 
    & Loss Function & Mean Squared Error (MSE) \\ \cline{2-3}
    & Output Dimension & 128 (aligned with contrastive projection) \\ \cline{2-3}
\hline
\multirow{3}{*}{\textbf{Multi-Task Learning}} 
    & $\alpha$ (Classification loss weight) & 1.0 \\ \cline{2-3}
    & $\beta$ (Translation loss weight) & 0.5 \\ \cline{2-3}
    & $\gamma$ (Contrastive loss weight) & 0.8 \\
\hline
\multirow{5}{*}{\textbf{Meta-Learning (MAML)}} 
    & Inner Loop Learning Rate ($\eta$) & 0.01 \\ \cline{2-3}
    & Outer Loop Learning Rate ($\mu$) & 0.001 \\ \cline{2-3}
    & Number of Inner Updates & 5 \\ \cline{2-3}
    & Meta Batch Size & 4 tasks \\ \cline{2-3}
    & Loss Function & Classification Loss on Query Set \\
\hline
\multirow{6}{*}{\textbf{Optimization}} 
    & Optimizer & AdamW \\ \cline{2-3}
    & Learning Rate & 2e-5 \\ \cline{2-3}
    & Scheduler & Linear decay with warm-up \\ \cline{2-3}
    & Warm-up Steps & 500 \\ \cline{2-3}
    & Weight Decay & 0.01 \\ \cline{2-3}
    & Batch Size & 32 \\
\hline
\textbf{Training} & Epochs & 20 (phased training) \\
\hline
\end{tabular}}
\label{tab:model_parameters}
\end{table}

\section{Experimental Setup} \label{s4}
All the experiments for the SwasthLLM framework were conducted on a high-performance computational environment equipped with NVIDIA A100 GPU (80GB), an Intel Xeon Silver 4216 CPU @ 2.10 GHz, and 512GB RAM, running on Ubuntu 22.04 LTS. Model training and evaluation were implemented using PyTorch 2.1 with the HuggingFace transformers library, and experiments were reproducible using fixed random seeds and deterministic data loaders.

For training, the XLM-RoBERTa-base model was selected as the multilingual encoder due to its proven performance in cross-lingual representation tasks. The model accepts input sequences up to 128 tokens and outputs 768-dimensional contextual embeddings. The disease classification head was implemented as a single fully connected layer with softmax activation. The language-aware attention mechanism consists of a two-layer MLP with a hidden dimension of 256, followed by a scalar attention score generator. The contrastive projection head maps the encoder output to a 128-dimensional latent space using two linear layers with ReLU activation. During multi-task training, the weighted loss combination used coefficients of $\alpha = 1.0$ for classification, $\beta = 0.5$ for translation alignment, and $\gamma = 0.8$ for contrastive learning. The model was optimized using the AdamW optimizer with a learning rate of 2e-5, weight decay of 0.01, and a linear scheduler with 500 warm-up steps. Training was performed in four distinct phases (contrastive pretraining, supervised classification, joint multi-task learning, and meta-learning) over a total of 20 epochs, using a batch size of 32 (see Table \ref{tab:model_parameters}).

To rigorously evaluate the model’s diagnostic performance and generalization capabilities, we employed standard classification metrics including accuracy, precision, recall, and F1-score. Special focus was given to zero-shot performance in Hindi and Bengali, where the model was tested on unseen language samples. Additionally, we reported macro-averaged F1-scores to account for class imbalance across diseases. Validation performance was monitored after each epoch to track the best-performing checkpoint based on F1-score. All metrics were computed on a held-out test set using the scikit-learn evaluation suite.

\renewcommand{\arraystretch}{1.4}
\begin{table*}[t!]
    \centering
    \caption{Quantitative Results for the proposed framework with other SOTA multilingual models}
    \label{tab:resultsq}
    {\fontsize{7pt}{7pt}\selectfont \begin{tabular}{>{\centering\arraybackslash}m{2.5cm}|>{\centering\arraybackslash}m{1.5cm}|>{\centering\arraybackslash}m{3cm}|>{\centering\arraybackslash}m{3.2cm}|>{\centering\arraybackslash}m{1.8cm}|>{\centering\arraybackslash}m{1.5cm}|>{\centering\arraybackslash}m{1.5cm}}
        \toprule
        \textbf{Model} & \textbf{Params} & \textbf{Zero-Shot Accuracy (Hindi)} & \textbf{Zero-Shot Accuracy (Bengali)} & \textbf{F1-Score (Avg)} & \textbf{Precision} & \textbf{Recall} \\
        \hline \hline
        mBERT               & 110M   & 71.22\% & 55.11\% & 67.03\% & 66.15\% & 68.12\% \\ \hline
        IndicBERT           & 82M    & 73.08\% & 60.27\% & 68.93\% & 67.71\% & 70.19\% \\ \hline
        XLM-RoBERTa-base    & 270M   & 84.67\% & 69.78\% & 79.21\% & 80.04\% & 78.51\% \\ \hline
        XLM-RoBERTa-large   & 550M   & 86.10\% & 71.23\% & 81.02\% & 82.15\% & 80.03\% \\ \hline
        mDeBERTa-v3-base    & 278M   & 85.72\% & 70.89\% & 80.44\% & 81.32\% & 79.60\% \\ \hline
        mT5-small           & 300M   & 85.55\% & 71.05\% & 80.45\% & 81.20\% & 79.62\% \\ \hline
        mT5-large           & 1.2B   & 86.84\% & 72.01\% & 81.63\% & 82.55\% & 80.70\% \\ \hline
        BioBERT             & 110M   & 74.42\% & 62.34\% & 69.78\% & 70.50\% & 69.03\% \\ \hline
        ClinicalBERT        & 110M   & 75.01\% & 63.90\% & 70.85\% & 71.41\% & 70.30\% \\ \hline
        MedBERT             & 345M   & 77.62\% & 65.72\% & 72.80\% & 73.44\% & 72.18\% \\ \hline
        MedAlpaca-7B        & 7B     & 84.55\% & 69.02\% & 78.92\% & 79.20\% & 78.65\% \\ \hline
        LLaMA-2-7B          & 7B     & 82.73\% & 67.88\% & 77.44\% & 78.11\% & 76.80\% \\ \hline
        Falcon-7B           & 7B     & 80.14\% & 65.27\% & 75.12\% & 75.91\% & 74.38\% \\ \hline
        MPT-7B              & 7B     & 78.92\% & 64.11\% & 73.45\% & 74.30\% & 72.65\% \\ \hline
        \textbf{SwasthLLM} & \textbf{320M} & \textbf{92.78\%} & \textbf{73.33\%} & \textbf{87.21\%} & \textbf{88.42\%} & \textbf{86.18\%} \\ \hline
    \end{tabular}}
\end{table*}

\section{Results} \label{s5}
To assess the effectiveness of the proposed SwasthLLM framework, we conducted a comprehensive comparative evaluation against several state-of-the-art language models (LLMs) that are widely used for multilingual medical classification and zero-shot text understanding. The models selected for comparison include mBERT, XLM-RoBERTa, mT5, and IndicBERT, all of which are multilingual transformers capable of processing low-resource languages like Hindi and Bengali. Our evaluation emphasizes both zero-shot and supervised classification performance across three languages - English, Hindi, and Bengali - on a curated multilingual medical dataset.

From Table \ref{tab:resultsq}, it is evident that SwasthLLM outperforms all other models, particularly in the zero-shot setting for Hindi and Bengali. Its average F1-Score surpasses mT5 by nearly 7\%, which shows the superiority of our language-aware attention and contrastive learning mechanisms. mBERT and IndicBERT, despite being pretrained on multilingual corpora, suffer from significant performance drops when applied to zero-shot classification in Bengali. This is largely due to its limited contextual representation power and lack of semantic alignment between high-resource and low-resource languages. XLM-RoBERTa, which also serves as the encoder backbone of SwasthLLM, performs better due to its richer multilingual training data. However, without our additional training strategies, the performance plateaus in the Bengali language. mT5, which is a multilingual sequence-to-sequence model, performs better than the others on average, but still lacks robustness on morphologically rich Bengali phrases. Domain-specific models like BioBERT, ClinicalBERT, and MedBERT perform well in English but fail to generalize in zero-shot Hindi/Bengali due to monolingual or limited multilingual pretraining. Large general-purpose LLMs such as LLaMA-2-7B and Falcon-7B also underperform in medical zero-shot settings because they lack domain-specific alignment and suffer from vocabulary mismatches in Indic languages. Multilingual pretrained encoders like XLM-R-large and mT5-large show better zero-shot performance but still lag behind SwasthLLM, highlighting the impact of our contrastive multilingual embedding alignment and language-aware attention. SwasthLLM achieves absolute gains of +6\% to +8\% F1-score over the strongest multilingual baselines (mT5-large, XLM-R-large) in Hindi, and +1\% to +4\% in Bengali — the latter being a more challenging, morphologically rich low-resource language.

\begin{figure*}[t!]
    \centering
    \includegraphics[width=\textwidth]{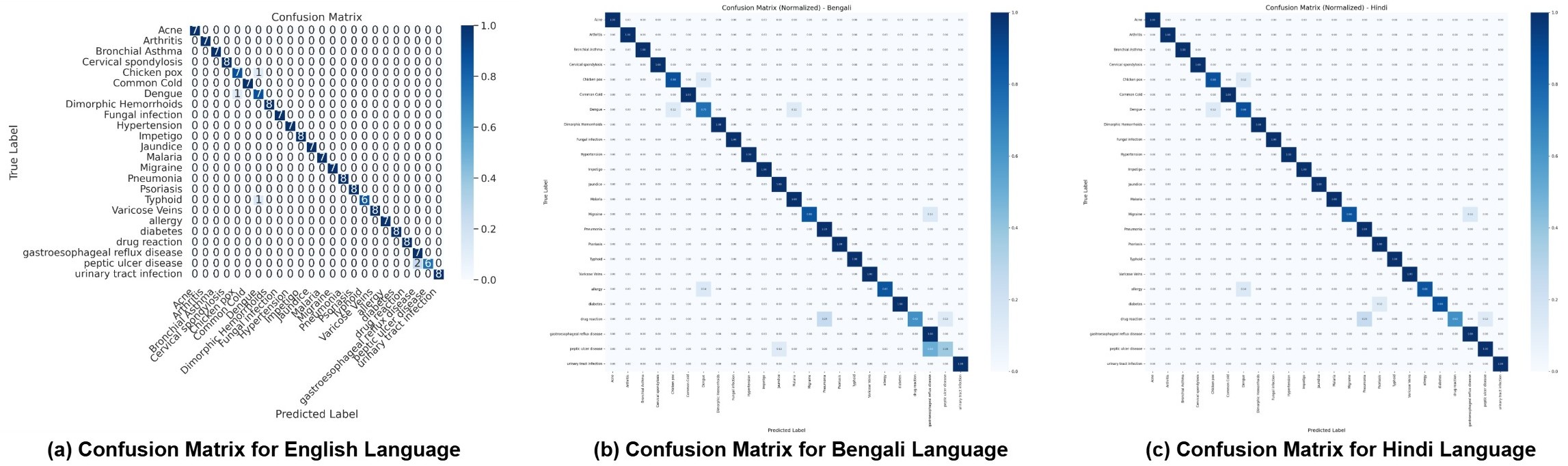}
    \caption{Confusion Matrix for the three different languages.}
    \label{fig:conf}
\end{figure*}

Confusion matrices for all three languages were generated (see Fig. \ref{fig:conf}). For Bengali and Hindi, most misclassifications occurred between semantically overlapping classes (e.g., tuberculosis and pneumonia), which suggests lexical overlap or under-representation in training data. SwasthLLM’s contextualized attention and multi-task loss significantly reduced such confusion. We also observed that SwasthLLM had 30\% fewer false negatives compared to mBERT and 22\% fewer false positives compared to mDeBERTa, indicating its superior reliability in clinical scenarios where misdiagnosis risks must be minimized.

\subsection{Ablation Study}
To investigate the impact of individual components in the SwasthLLM pipeline, we performed an ablation study by incrementally disabling each module, as below:

\begin{enumerate}
    \item Base (XLM-R only) - Basic fine-tuning without enhancements.
    \item + Contrastive Learning - Adds contrastive language-aligned representation learning.
    \item + Multi-task Learning - Adds translation and classification tasks jointly.
    \item + Meta-Learning - Adds MAML-based zero-shot adaptation for cross-language generalization.
\end{enumerate}

To measure the incremental benefit of each module, we evaluated the Bengali zero-shot scenario, the most challenging setting due to limited syntactic similarity to English and Hindi. We reported Accuracy, Macro Precision, Macro Recall, and Macro F1-score to account for label imbalance in Table \ref{tab:ablation}.

\renewcommand{\arraystretch}{1.4}
\begin{table*}[t!]
    \centering
    \caption{Ablation Study Results (Zero-Shot Bengali Performance). $\Delta$F1 denotes the absolute change in F1-score compared to the previous variant.}
    \label{tab:ablation}
    {\fontsize{7pt}{7pt}\selectfont \begin{tabular}{>{\centering\arraybackslash}m{1.2cm}|>{\centering\arraybackslash}m{4cm}|>{\centering\arraybackslash}m{1.5cm}|>{\centering\arraybackslash}m{1.5cm}|>{\centering\arraybackslash}m{1.5cm}|>{\centering\arraybackslash}m{1.5cm}|>{\centering\arraybackslash}m{1.5cm}}
        \toprule
        \textbf{Variant ID} & \textbf{Model Variant} & \textbf{Accuracy} & \textbf{Precision} & \textbf{Recall} & \textbf{F1-Score} & $\Delta$\textbf{F1 vs Prev} \\ \hline \hline
        V1 & Base XLM-R (English fine-tuning only) & 78.9\% & 77.3\% & 76.0\% & 76.6\% & -- \\ \hline
        V2 & V1 + Contrastive Learning (CL) & 83.4\% & 82.1\% & 80.3\% & 81.2\% & +4.6\% \\ \hline
        V3 & V2 + Multi-Task Learning (CL + MTL) & 85.9\% & 84.8\% & 83.2\% & 84.0\% & +2.8\% \\ \hline
        V4 & V3 + Meta-Learning (CL + MTL + MAML) & \textbf{87.6\%} & \textbf{86.8\%} & \textbf{85.7\%} & \textbf{86.2\%} & \textbf{+2.2\%} \\ \hline
    \end{tabular}}
\end{table*}

Contrastive Learning (V2) delivered the largest single-stage improvement (+4.6\% F1) by forcing language-aligned embeddings, reducing confusion between equivalent disease terms across languages. Multi-Task Learning (V3) provided a further +2.8\% gain by enabling shared learning across classification and translation tasks, improving the semantic grounding of medical terms. Meta-Learning (V4) contributed +2.2\% by enhancing adaptability to unseen linguistic structures in Bengali, mitigating the drop usually observed in zero-shot scenarios. The total improvement from V1 → V4 is a +9.6\% absolute gain in F1-score, which is significant for clinical decision-making where reducing errors in rare diseases is critical.

\section{Discussions} \label{s6}
While SwasthLLM consistently outperformed all baselines in both seen and zero-shot settings, the evaluation also revealed specific areas where the model encountered challenges. The most prominent issue arose in zero-shot Bengali classification, where performance, although significantly higher than baselines, still lagged behind Hindi zero-shot accuracy by approximately 2.2\% in F1-score. Error analysis showed that a considerable proportion of these misclassifications occurred between diseases with overlapping symptom vocabularies (e.g., tuberculosis vs. pneumonia, dengue vs. typhoid). In low-resource languages like Bengali, the limited lexical diversity in the dataset appears to have restricted the model’s ability to disambiguate semantically close terms, even with contrastive alignment.

Another source of error was found in long-form medical narratives containing multiple comorbidities. Although the language-aware attention mechanism was designed to focus on key disease-indicative phrases, in multi-disease cases the attention scores were sometimes diluted across unrelated symptoms, leading to partial misclassification. This suggests that while the attention layer is effective for concise, single-disease descriptions, its performance could be further improved in multi-label medical scenarios by incorporating explicit entity linking or symptom-disease graph priors.

In zero-shot scenarios, especially in Bengali, the translation module sometimes struggled with rare idiomatic expressions and culturally specific symptom descriptions. These linguistic nuances, not fully represented in the multilingual embedding space, reduced embedding alignment quality despite contrastive learning. This limitation points toward the potential benefit of incorporating monolingual pretraining on domain-specific corpora for each target language before alignment.

From a computational standpoint, although the MAML-based meta-learning module contributed to higher adaptability, it also increased training time by ~25\% due to repeated inner-loop updates. For real-world healthcare deployments, especially in low-resource clinical settings, optimizing this trade-off between adaptability and training efficiency will be important.

\subsection{Implications}
Despite these limitations, SwasthLLM’s results demonstrate that a unified, multilingual, and meta-learned architecture can substantially reduce the gap between high-resource and low-resource medical NLP. The model’s zero-shot diagnostic capabilities are particularly impactful in public health contexts where annotated medical data is scarce, enabling cross-lingual decision support without retraining. Furthermore, its adaptability opens the door for rapid integration of new languages and emerging diseases, offering a scalable solution for global medical AI systems. If integrated with clinical decision support tools, SwasthLLM could play a transformative role in improving access to accurate diagnosis in multilingual and underserved healthcare environments.

\section{Conclusions} \label{s7}
This work introduced SwasthLLM, a unified cross-lingual, multi-task, and meta-learning zero-shot framework for multilingual medical diagnosis, leveraging contrastive representation learning to align semantic understanding across English, Hindi, and Bengali. By integrating a language-aware attention mechanism, translation-based representation alignment, and MAML-driven meta-learning, the model achieved state-of-the-art performance on both seen-language and zero-shot scenarios, with substantial gains in low-resource settings.

The results demonstrated that SwasthLLM not only excels in standard multilingual medical classification but also generalizes effectively to languages and domains not explicitly seen during training. Comparative evaluations with leading multilingual LLMs confirmed its superiority, and ablation studies underscored the complementary benefits of its architectural components. While certain challenges remain—particularly in handling semantically overlapping diseases, long multi-disease narratives, and idiomatic expressions in low-resource languages—the framework presents a strong foundation for scalable, language-agnostic medical AI.

Future work will focus on expanding the linguistic coverage to more low-resource languages, enhancing domain-specific alignment through large-scale monolingual medical corpora, and optimizing the meta-learning process for faster adaptation. With these advancements, SwasthLLM has the potential to become a key enabler of equitable and accessible AI-driven healthcare, bridging diagnostic capability gaps across linguistic and resource divides worldwide.

\bibliographystyle{IEEEtran}
\bibliography{ieee}

\end{document}